\documentclass[3p,times,procedia]{elsarticle}
\usepackage{ecrc}
\usepackage{floatrow}
\newfloatcommand{capbtabbox}{table}[][\FBwidth]

\volume{00}

\firstpage{1}

\journalname{Transportation Research Procedia}

\runauth{Cabrejas-Egea A., Zhang, R., and Walton, N.}

\jid{trpro}

\usepackage{amssymb}

\biboptions{authoryear}

\usepackage[figuresright]{rotating}

\usepackage[bookmarks=false]{hyperref}
    \hypersetup{colorlinks,
      linkcolor=blue,
      citecolor=blue,
      urlcolor=blue}
\usepackage{natbib}
\usepackage{subcaption}

\begin{document}
\begin{frontmatter}



\dochead{14th Conference on Transport Engineering: 6th – 8th July 2021}%

\title{Reinforcement Learning for Traffic Signal Control: Comparison with Commercial Systems}


%

\author[a,b]{Alvaro Cabrejas-Egea\corref{cor1}} 
\author[a,c]{Raymond Zhang}
\author[a,d]{Neil Walton}

\address[a]{The Alan Turing Institute, 96 Euston Rd, London NW1 2DB, United Kingdom}
\address[b]{MathSys CDT, University of Warwick, Coventry CV4 7AL, United Kingdom}
\address[c]{Ecole Normale Superieure de Paris-Saclay, 4 Avenue des Sciences, 91190 Gif-sur-Yvette, France}
\address[d]{University of Manchester, Oxford Rd, Manchester M13 9PL, United Kingdom}

\begin{abstract}
Recently, Intelligent Transportation Systems are leveraging the power of increased sensory coverage and computing power to deliver data-intensive solutions achieving higher levels of performance than traditional systems. 
Within Traffic Signal Control (TSC), this has allowed the emergence of Machine Learning (ML) based systems. Among this group, Reinforcement Learning (RL) approaches have performed particularly well. 
Given the lack of industry standards in ML for TSC, literature exploring RL often lacks comparison against commercially available systems and straightforward formulations of how the agents operate. 
Here we attempt to bridge that gap.
We propose three different architectures for TSC RL agents and compare them against the currently used commercial systems MOVA, SurTrac and Cyclic controllers and provide pseudo-code for them.
The agents use variations of Deep Q-Learning and Actor Critic, using states and rewards based on queue lengths.
Their performance is compared in across different map scenarios with variable demand, assessing them in terms of the global delay and average queue length.
We find that the RL-based systems can significantly and consistently achieve lower delays when compared with existing commercial systems. 

\end{abstract}

\cortext[cor1]{Corresponding author. Tel.: +44-7511-398-261.}
\end{frontmatter}

\email{acabrejasegea@turing.ac.uk - Pre-print version, submitted to CIT2021.}



\graphicspath{{figures/}}
\vspace{-5mm}
\section{Introduction}
Traffic Signal Control (TSC) can be used to ensure the safe and efficient utilisation of the road network at junctions, where traffic can change directions and merge, having to manage conflicting individual priorities with the global needs of the network.
Traffic congestion has a major financial impact. A study by \cite{inrix_scorecard_nodate} shows that traffic congestion in 2019 cost £$6.9$ billion in the UK alone, with similar patterns being observed in other developed countries.
Cities around the globe are starting to explore the deployment of smart Urban Traffic Controllers (UTCs) that use real time data to adjust their stage schedule and green time duration.
Traditionally, fixed time plans have been used, with systems that optimise the green time splits in a deterministic manner, requiring costly site-specific knowledge and typical demand profiles to provide effective control.
These methods are not easily scalable and deteriorate over time as the traffic demand changes (\cite{bell1986ageing}).
With the development of induction loops, real time actuated UTCs were created in two variants: those that optimise single isolated intersections with systems such as MOVA (\cite{MOVA}), and those that cover multiple intersections such as SCOOT (\cite{SCOOT}).
To remedy the scalability problem, other systems are based on local rules, generating a self-organising area traffic controllers, such as SurTrac (\cite{SURTRAC}).
With the breakthrough of Deep Reinforcement Learning (DRL) on complex problems such as Atari games or Go (\cite{mnih2013playing, silver2017mastering}), attention has turned towards adapting these approaches to generate industry-grade controllers for traditionally noisy and systems such as TSC.

This paper aims to reproduce some of the results of the main and most successful RL approaches on intersections of increasing complexity, while comparing different architectures of DRL TSC agents, since, given the complexity of their implementation, most available literature only deals with a single class.
\vspace{-3mm}
\section{State of the Art}
\vspace{-1mm}
\label{rlref}
\noindent We discuss the state of the art in reinforcement learning in signal control. We discuss the state of the art systems for commercial simulation and traffic signal control which we analyse in this study.
\vspace{-1mm}
\subsection{Previous Work} 
\vspace{-1mm}

RL is an area of ML aiming to imitate how biological entities learn, where an agent evolves in an unknown environment, learning how to perform with no prior information, based on its interactions said environment.
The agent aims to maximise a reward signal it receives as a feedback for its actions.
RL methods have been applied to TSC in experimental setups.
While there is a variety of approaches in the literature that craft successful RL-based TSC systems, most do no present comparisons against commercial systems that are the concern of this paper.

Recent works \cite{gao2017,wan2018,mousavi2017} use neural networks as function approximators to avoid the dimensionality and computing limitations of table based methods in large state-action spaces, showing DRL TSC can be more efficient than some earlier methods.
The first two use discreet cell encoding vectors to represent the system, which are passed to a Convolutional Neural Network (CNN), whereas the second directly uses pixels in the same manner. 
\cite{gao2017} compared the results against a fixed time and longest-queue-first systems, finding RL to perform better, while \cite{wan2018} found similar results comparing against a fixed time system.
\citet{liang2018}, used the same initial approach and compared against two different fixed-time systems, ranking better than both and providing some early evidence of the benefits of using Double DQN (\cite{hasselt2010double}), Duelling architecture (\cite{wang2015dueling}) and Prioritised Experience Replay (PER) (\cite{schaul2015prioritized}).
\citet{genders2018} evaluated different state representations, finding little difference in the performance of the agents as a result of the change in the magnitudes observed.
In \citet{stevanovic2008split} the authors compare SCOOT with a Genetic Algorithm-based control method. It is shown that SCOOT's performance can be surpassed by more adaptive Genetic Algorithms that, in turn, tend to be less effective at learning than RL methods.

Despite these previous works, most results are hard or impossible to reproduce given the lack of industry standards in terms of simulators, performance metrics, the lack of availability of commercial algorithms for comparison and the fierce protection of their internal workings, and the lack of open-source code of proposed RL models.
\vspace{-1mm}
\subsection{Commercial Traffic Signal Control Simulators and Optimisers}
\vspace{-1mm}

\noindent \textbf{PTV Vissim.} Our simulations are conducted on PTV Vissim, a state-of-the-art commerical traffic simulator that is able to produce a wide variety of traffic demands over an array of signal controls and road traffic scenarios. We interface our RL algorithms via COM interface, allowing the interface with Tensorflow to construct deep learning agents. 

\smallskip
\noindent \textbf{MOVA} (Microprocessor Optimised Vehicle Actuation, \cite{MOVA}) is a traffic controller designed by TRL Software. It aims to reduce delay on isolated junctions. 
The basic functioning of MOVA involves two induction loop detectors estimating the flow of vehicles in each lane.
The system makes a virtual cell representation of the lanes within MOVA, and then it computes a performance index based on the delays calculated.
If the index results lower than a certain threshold, the signal is changed to the next stage, otherwise the stage it is extended. 

\smallskip
\noindent \textbf{Surtrac} (Scalable URban TRaffic Control, \cite{SURTRAC}) is decentralised, with each intersection allocating green time independently and asynchronously based on incoming flows.
Each intersection is controlled by a local scheduler and communicates projected outflows to the downstream neighbouring junctions, modelling vehicles as a sequence of clusters.
This allows for locally balancing competing flows while creating "green corridors" by finding an optimal sequence such that the input jobs (ordered clusters) are cleared while minimising the joint waiting time.
\vspace{-1mm}
\section{Methods}
\vspace{-1mm}
\label{sec:methods}
\vspace{-1mm}
\noindent \textbf{Traffic Control as a Markov Decision Process.}
\label{sec:mdp}
The control problem can be formulated as a Markov Decision Process (MDP) defined in terms of a 5-tuple: A set of possible environment states $s\in\mathcal{S}$, a set of actions of the agent $a\in\mathcal{A}$, a stochastic transition function $\forall a \in \mathcal{A},  \mathcal{T}^a_{s,s'}  \triangleq \mathbb{P}(s_{t+1}=s'| s_t=s,a_t=a )$, a scalar real valued reward function $R(s_t,s_{t+1},a_t)$ providing a performance measure to the transition generated by progressing into the state $s_{t+1}$ after taking action $a_t$ while in state $s_t$, and a discount factor $\gamma$ that will provide the balance between immediate exploitation and approaches that aim to maximise returns over time.
In the case of TSC, the MDP is modelled as partially observable, following an unknown stochastic transition function.

\medskip
\noindent \textbf{Reinforcement Learning.}
The goal of the agents is to maximise their future discounted return, $ G_t = \sum_{t=0}^{+\infty} \gamma^t r(s_t,a_t) $ with $\gamma$ $ \in [0,1] $ by learning a policy $\pi$, parametrised by the weights $\theta$ of the neural network performing the approximation of the reward function and mapping states to actions: $\pi:\mathcal S\rightarrow \mathcal A$.
The reward function maps an action given a state to a reward scalar value: $r: \mathcal S \times \mathcal A \rightarrow \mathbb{R}$.
The action-value function or Q-value is $ Q_\pi(s,a) = \mathbb{E}_{\pi}[G_t| s_t =s, a_t=a ] $ representing the total episodic return by following $\pi$ after being in state $s$ and taking action $a$. 

\medskip
\noindent \textbf{Value-based Reinforcement Learning Methods - DQN.}
Tabular value-based methods, such as Q-Learning, attempt to learn an optimal policy $Q_{\pi}^*=\max_{\pi} \mathbb{E}[r_t|s_t=s,a_t=a]$ by iteratively performing Bellman updates on the Q-values of the individual state-action pairs:
\begin{equation}
Q_{\pi}(s_t,a_t)\leftarrow Q_{\pi}(s_t,a_t) + \alpha \big( y_t - Q_{\pi}(s_{t+1},a_{t+1}) \big),\,\,\,\, \textrm{with}\,\,\,\, y_t = r_t + \gamma \max_{a'} Q_{\pi} (s_{t+1},a',\theta'),
\label{eq:valuebasedupdate}
\end{equation}
where $\alpha$ is the learning rate and $y_t$ is the Temporal Difference (TD) target for the value function.

Deep Q-Network (DQN) agents are an evolution of Q-Learning.
The purpose of the agent is to find an approximation of $Q_{\pi}*$ by tuning the weights $\theta$ of a neural network. 
The agent keeps a second neural network, the target network, parametrised by the weights vector $\theta'$ which is used to generate the TD targets.

\begin{figure}
\begin{subfigure}{.45\textwidth}
  \centering
  \includegraphics[width=.95\linewidth]{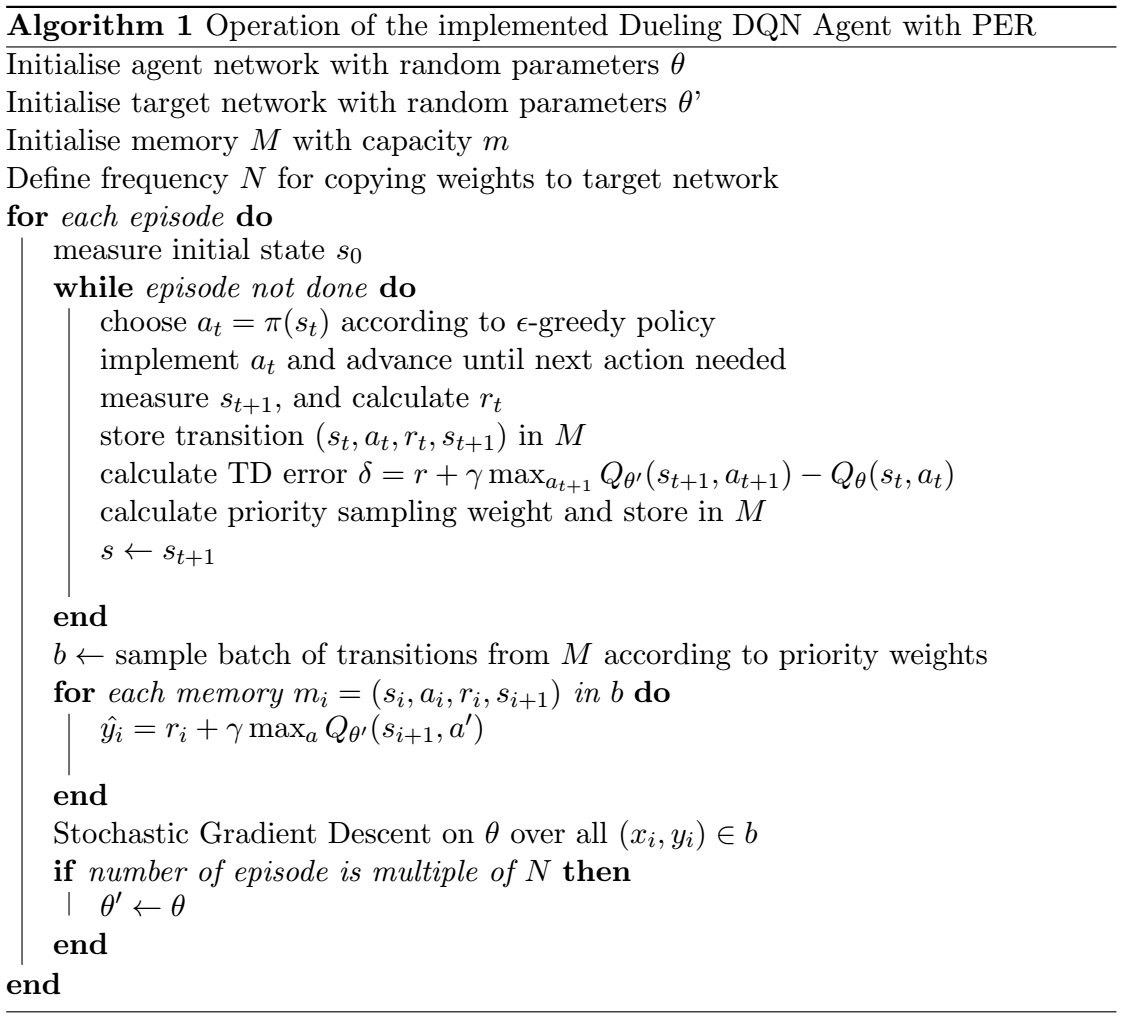}
  \caption{}
  \label{fig:DQNalgo}
\end{subfigure}%
\begin{subfigure}{.45\textwidth}
  \centering
  \includegraphics[width=.95\linewidth]{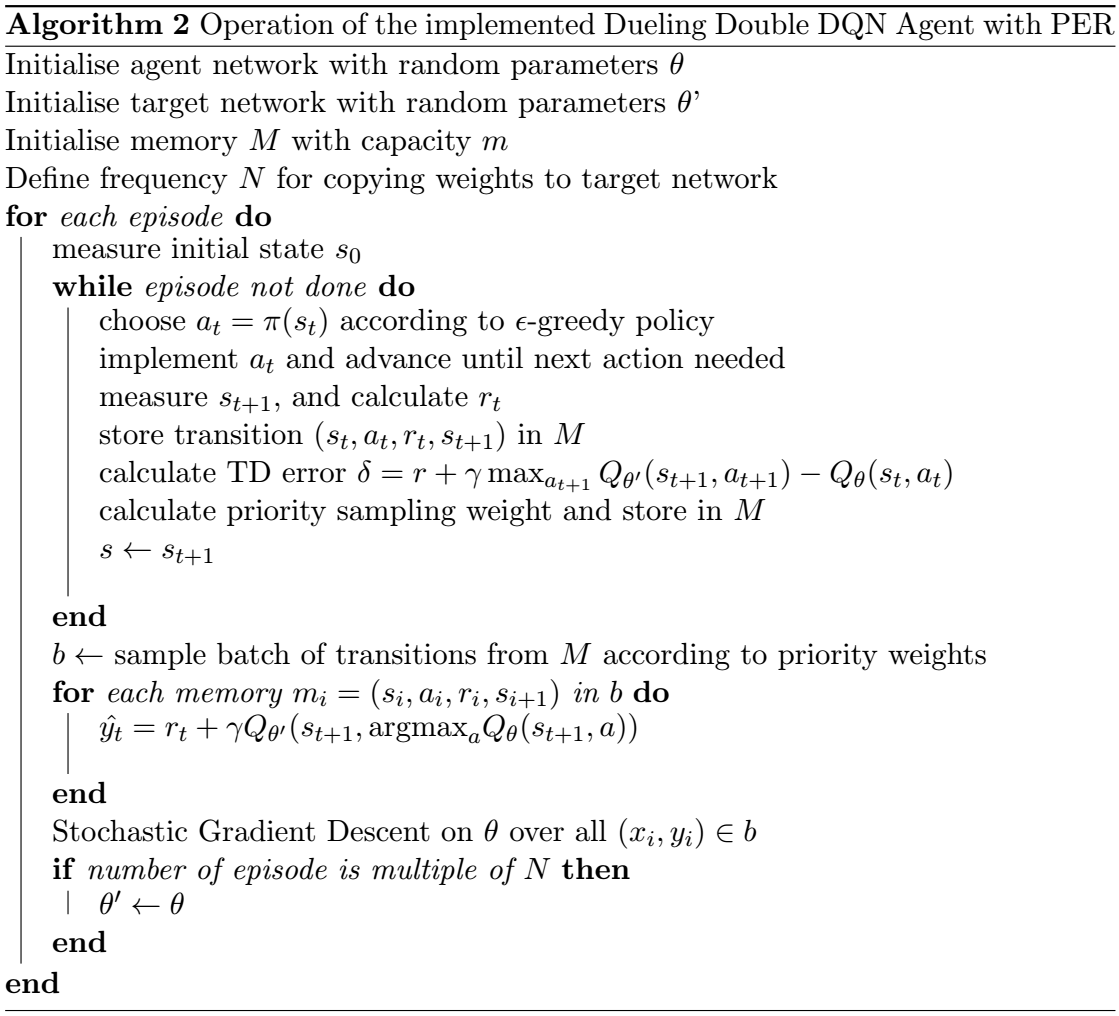}
  \caption{}
  \label{fig:DDQNalgo}
\end{subfigure}
\caption{(a): DDQN Pseudocode. (b): DDDQN Pseudocode.}
\label{fig:DQNPseudo}
\end{figure}

\begin{figure}
\begin{subfigure}{.45\textwidth}
  \centering
  \includegraphics[width=.95\linewidth]{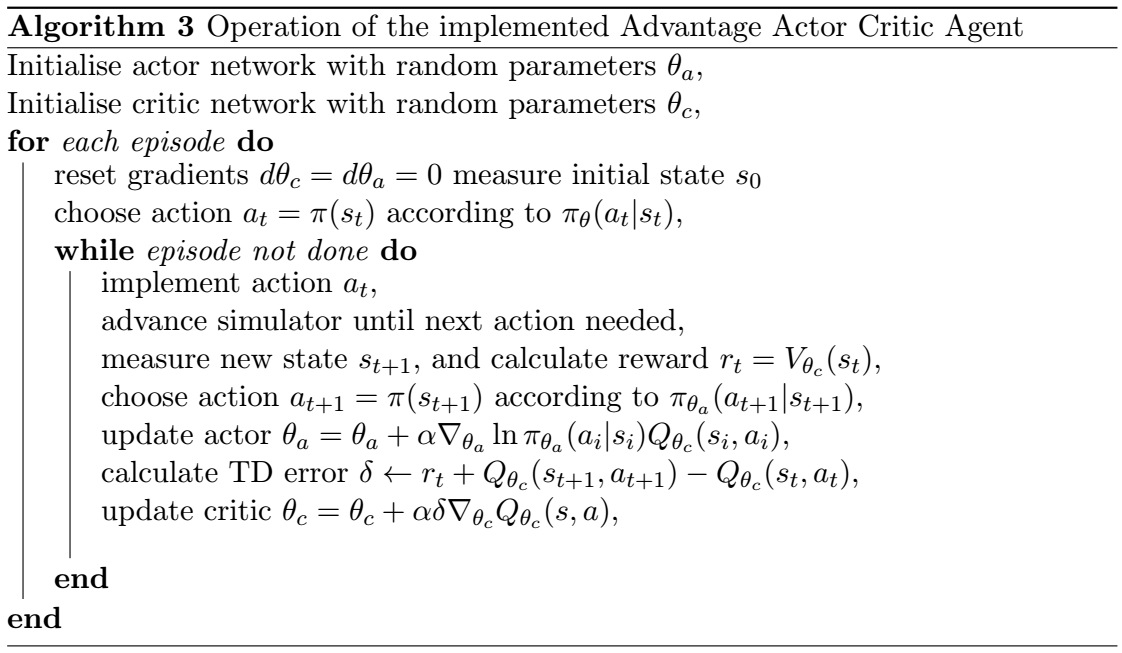}
  \caption{}
  \label{fig:a2calgo}
\end{subfigure}%
\begin{subfigure}{.45\textwidth}
  \begin{subfigure}{.5\textwidth}
  \centering
  \includegraphics[width=.77\linewidth]{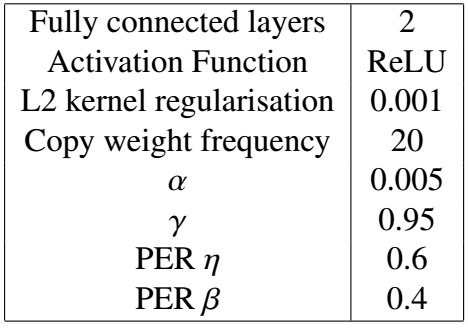}
  \caption{}
  \label{fig:DQNhyp}
\end{subfigure}%
\begin{subfigure}{.5\textwidth}
  \centering
  \includegraphics[width=.95\linewidth]{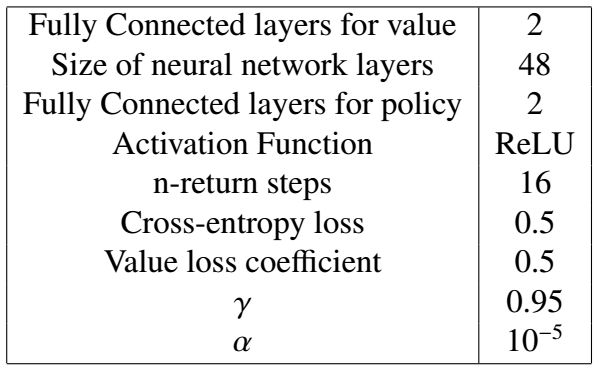}
  \caption{}
  \label{fig:A2Chyp}
\end{subfigure}
    
\end{subfigure}
\caption{(a): A2C Pseudocode. (b): DDQN/DDDQN Hyperparameters. (c): A2C Hyperparameters.}
\label{fig:A2CPseudo}
\end{figure}

The experience replay memory increases training stability, obtaining samples that cover a wider range of situations and that can be used several times for gradient descent.
Three additional modules have been applied to the agent to improve performance, Double Q Learning, PER, and Dueling Architecture.
The DQN variants implemented are described on the algorithms displayed in Figs. \ref{fig:DQNalgo} and \ref{fig:DDQNalgo}, and use the hyperparameters described in Fig. \ref{fig:DQNhyp}.

\medskip
\noindent \textbf{Policy Gradient Reinforcement Learning Methods - A2C. }
Policy Gradient in RL is based on the idea that obtaining a direct policy $\pi(s)$ mapping states to actions can be easier than estimating the value function or the state-action values.
It has an added benefit in that it can learn stochastic policies, generating a probability distribution over the potential actions.
The goal is to find the policy that maximises the reward. To do so one has to perform gradient ascent on the performance measure $J = \sum_{a\in \mathcal A} \mathbb E [Q(s_0,a) \pi(a|s)]$.

The Advantage Actor Critic (A2C) method tries to reduce the variance in the policy method by combining the direct mapping from actions with the value-based approximation method.
The goal is to learn an actor 
\begin{equation}
\pi_{\theta} = \mathbb{P}_{\theta}[a_t = a|s_t = s], \,\,\,\,\textrm{and a critic}\,\,\,\,  V_{\pi}^{\theta}(s) = \mathbb{E}_{\theta}[G_t|s_t=s],
\label{A2C:Actor}
\end{equation}
both of which are parametrised by the neural network weights vector $\theta$.
The pseudocode for the A2C agent can be found in Fig.\ref{fig:a2calgo} and its hyperparameters are displayed in Fig.\ref{fig:A2Chyp}.

\vspace{-1mm}
\subsection{State, Action and Rewards of the agents}
\label{slap}
\vspace{-1mm}
\noindent \textbf{Agent Configuration. }
All the experiments here presented use the same descriptions for state and reward calculation, differing in the number of actions available to them.
The state of an intersection of $l$ lanes will be presented to the agents as a state vector $s\in\mathbb{R}
^{l+1}$, in which each component represents the length of the queue of vehicles measured upstream from the traffic light in metres.
The last component will be current stage being implemented.

While marginal improvements in performance can be obtained by using different variables for reward (\cite{egearlutc,multimodalarxiv}), as per the discussion of \citet{heydecker2004}, queues can be a reasonable choice for states and rewards, being able to transmit useful information to the agent relative to the mean rate of delay of the system.
Based on this, the reward after an action will be calculated as the negative sum of the length of the queues of all lanes immediately upstream from the intersection: $r_t  = - \sum_l q_{l,t}.$

The agents were trained using a fixed vehicle demand of $400$ vehicles per hour on each of the incoming lanes.
Both DQN variants were trained for $400$ episodes, using an $\epsilon$ geometrically annealed from $1$ to $0.001$.
The A2C agents were trained for $100$ episodes until they converged. 
Best performing agents in each class were selected for benchmarking and evaluated in scenarios lasting one hour.

\vspace{-1mm}

\medskip
\noindent \textbf{Agent Benchmarking. }
\label{bench}
In order to compare the agents performance, a testing framework was defined. For each model, a demand profile will be created, following the shape found in a \emph{typical day} using the methodology introduced in \cite{egea2018estimating} and expanded in \cite{ieeewarp}.
The profile will be split on 10 segments of length 6 minutes.
Each of these will correspond with a level of demand.
The demand levels are obtained by setting the maximum demand the intersection will suffer, setting that magnitude to coincide with the peaks of the distribution that could be found on said \emph{typical day}, and will be specified in each experiment's section.
Random seeds are updated after every simulation episode, training or testing.
The quantitative metrics on which the system will be evaluated are the Global Cumulative Delay (deviations from free-flow time) and the Average Queue Length generated during the evaluation.
\vspace{-4mm}
\section{Experimental Results}
\vspace{-2mm}

\label{experiments}
\begin{figure}
\begin{subfigure}{.3\textwidth}
  \centering
  \includegraphics[width=.95\linewidth]{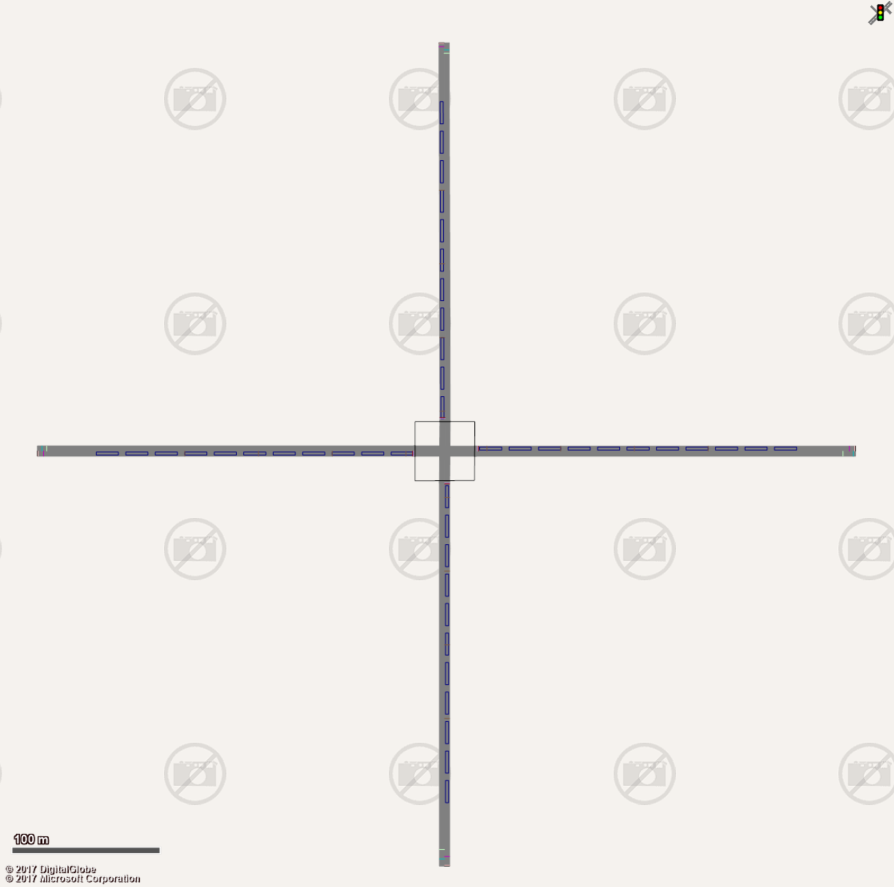}
  \caption{}
  \label{fig:sfig1}
\end{subfigure}%
\begin{subfigure}{.3\textwidth}
  \centering
  \includegraphics[width=.95\linewidth]{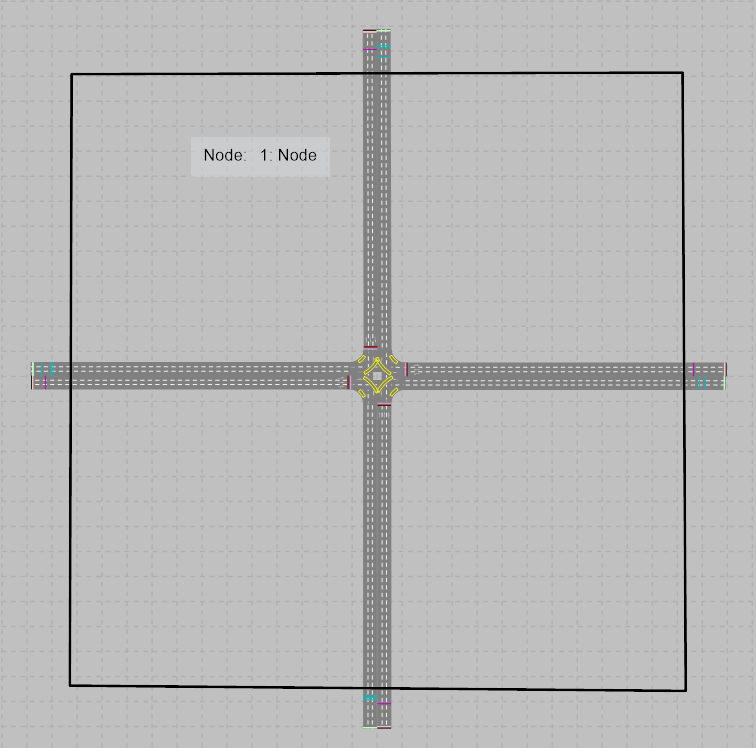}
  \caption{}
  \label{fig:sfig2}
\end{subfigure}
\caption{(a): Cross Straight Map. (b): Cross Triple Map.}
\label{fig:Maps}
\end{figure}
\medskip
\noindent \textbf{Experiment 1: Cross Straight.}
\label{singlecrossstraight}
The first test is conducted on the simplest junction, shown in Fig \ref{fig:sfig1}.
The junction is composed on 4 lanes distributed in 4 arms.
The controller has two stages, a north-south stage and an east-west stage, and turning is not allowed. 
The aim was to perform an initial performance comparison of DRL algorithms against MOVA, SUTRAC, and a cyclic controller.
Here the goal for the agent was to exert fine adaptive timing control while extrapolating, rather than using complicated transitions between stages. 
MOVA was configured using loop detectors set in accordance to its manual, the implementation of Surtrac follows the work of \citet{Xie2012} .
During evaluation, an average of 2120 vehicles are inserted in the model, with 2 peaks of demand of 3000 veh/h for 6 minutes each.

Figure \ref{fig:scsresu} and Table \ref{table:queuescs} show the Global Cumulative Delay and average queue length for the network.
As expected, the cyclic solution is outperformed by all adaptive controllers.
The different controllers are on a par with a slight advantage for the DuelingDDQN which saves the community an average of 3000 seconds compared to MOVA on this hour of simulation, which represents on average 1-2 seconds per vehicle.
RL agents also seem slightly more robust against changes in demand, producing lower slopes in the delay graphs in sections of extreme demand.

\begin{figure}
\begin{subfigure}{.5\textwidth}
  \centering
  \includegraphics[width=1.1\linewidth]{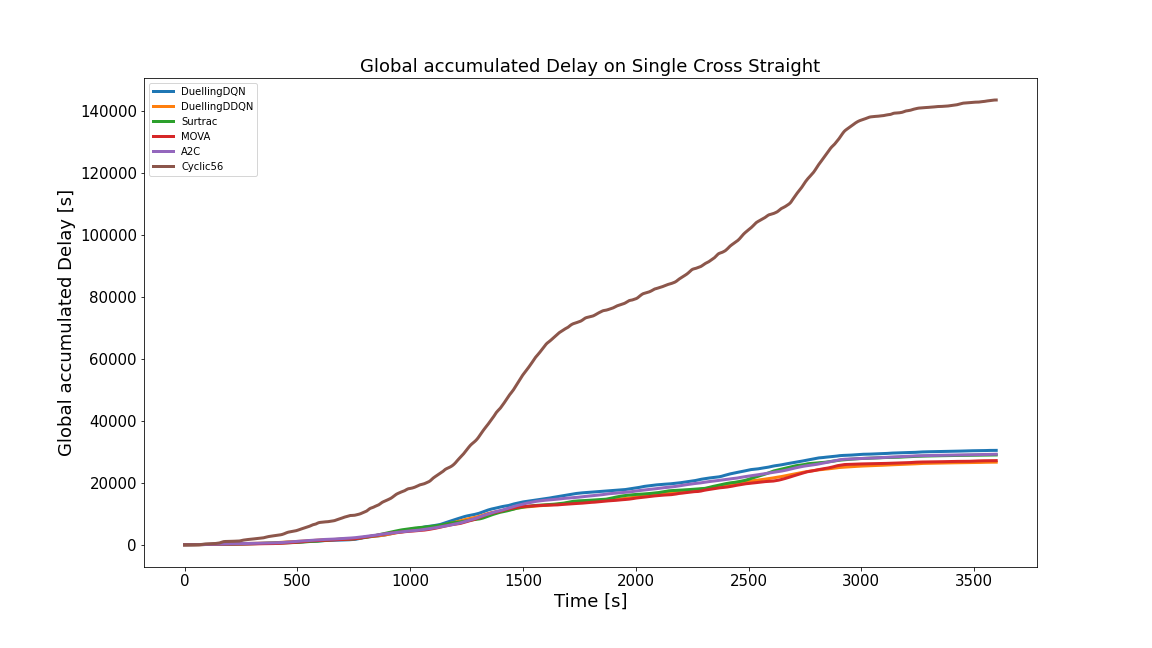}
  \caption{}
  \label{fig:scsdelay}
\end{subfigure}%
\begin{subfigure}{.5\textwidth}
  \centering
  \includegraphics[width=1.1\linewidth]{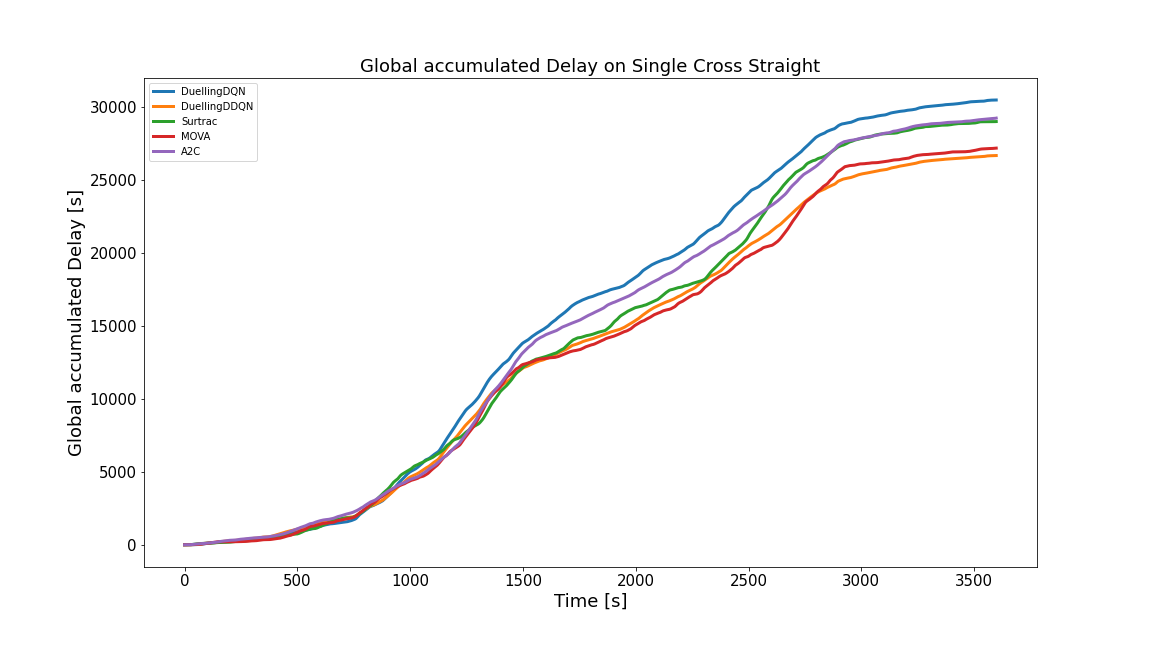}
  \caption{}
  \label{fig:scsdelaywo}
\end{subfigure}
\caption{Global Cumulative Delay in Single Cross for DQN variants, A2C, Surctrac, MOVA and the reference Cyclic controller.}
\label{fig:scsresu}
\end{figure}

\begin{table}[htbp]
	\caption{Cumulative Delay and Cumulative Stop Delay in seconds across Controllers on Single Cross Straight.}
	\begin{center}
		\begin{tabular}{|c|c|c|}
			\hline
			\textbf{Controller}&{\textbf{Cumulative Delay [s]}}&{\textbf{Average Sum of Queues [m]}} \\
			\hline
			Cyclic& 143660.50& 132.37\\
			\hline
			MOVA& 27187.53 & 60.59\\
			\hline
			SURTRAC& 29008.36 & 72.41\\
			\hline
			A2C& 26382.14 & 56.07\\
			\hline
			DDQN& 28303.94 & 50.11\\
			\hline
			DDDQN& 21286.86 & 49.42\\
			\hline
		\end{tabular}
		\label{table:queuescs}
	\end{center}
\end{table}

The cyclic controller resulted in saturated lanes during both peaks and queues in excess of $150$ metres during a great part of the simulation.
MOVA suffered two moments in which at least a sensor was saturated coinciding with the peaks in demand, however the queues were close to lengths of around $50$ metres during the most part of the simulator.
Surtrac followed a similar pattern, having a single lane saturated coinciding with the second peak in demand.
RL agents as suffered no saturation in any of their lanes during the length of the evaluation. 
They all managed a more balanced distribution of queues in their respective lanes, displaying a higher ability to balance loads during peak times.
Because of the simplicity of this 2 actions intersection, there is not a lot of delay difference between adaptive UTCs.
As it will be appreciated shortly, these results will change when we consider more complex junctions.

Given the difference in performance between the adaptive and cyclic controllers, which is expected to become greater on more complex intersections, and the increasing difficulty in setting them in  large intersections, the cyclic controller will be omitted for the next examples.
Given that the A2C agent has been clearly outperformed in this experiment by those based on the DQN architecture, the following experiments will focus on the performance of this last architecture compared with commercial systems.
\vspace{-1mm}

\medskip
\noindent \textbf{Experiment 2: Cross Triple - 4 actions.}
\label{tripletrain4}
This junction, as shown in Fig. \ref{fig:sfig2}, displays a much higher complexity than the intersection presented in the previous section.
It is composed of 4 incoming links of 3 lanes each. 
In each incoming link, the left lane serves a dedicated nearside turning lane, the central allows for forward travel and the right lane allows for both offside turning and going straight.
To mitigate this, the first experiment was run with agents that would take $4$ queue inputs, plus the state of the traffic signal as state input.
The action set was consequently limited to $4$ different actions, being allowed only those that set to green the $3$ traffic lights serving the lanes of the same incoming link.
This allows for turning vehicles, but prevents more sophisticated stages from happening.
During the hour of evaluation, the demand profile from the last experiment was used with a scaling factor of $1.5$, an average of $3180$ vehicles were introduced to the model, with $2$ peaks of demand of $4500$ vehicles/hour for 6 minutes each.

As it can be seen in Table \ref{table:queuesct4} the UTC using MOVA performs poorly compared to the DQN-based agents.
During this hour of simulation RL agents halve the cumulative delay, saving over $27$ hours of travel time for all vehicles involved, an average of over $32$ seconds of per vehicle.
The length of the queues in those intersections controlled by RL agents during the test scenario were lower than the ones controlled by MOVA. Further in Figure \ref{fig:scsdelaywo} we see that DDQN  significantly outperforms MOVA over a wide range of loads and traffic scenarios.
It can be seen that the agent using Dueling Q-Learning has a better performance than that Dueling Double Q-Learning.
\begin{figure}
\begin{subfigure}{.5\textwidth}
  \centering
  \includegraphics[width=1.1\linewidth]{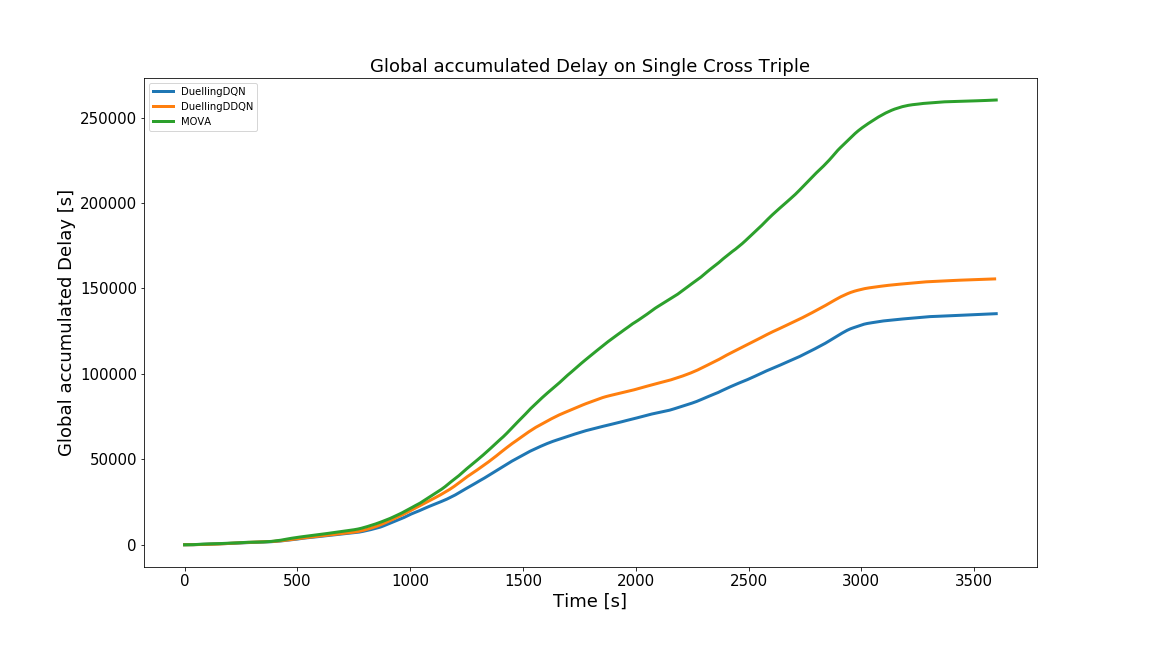}
  \caption{}
  \label{fig:scsdelay}
\end{subfigure}%
\begin{subfigure}{.5\textwidth}
  \centering
  \includegraphics[width=1.1\linewidth]{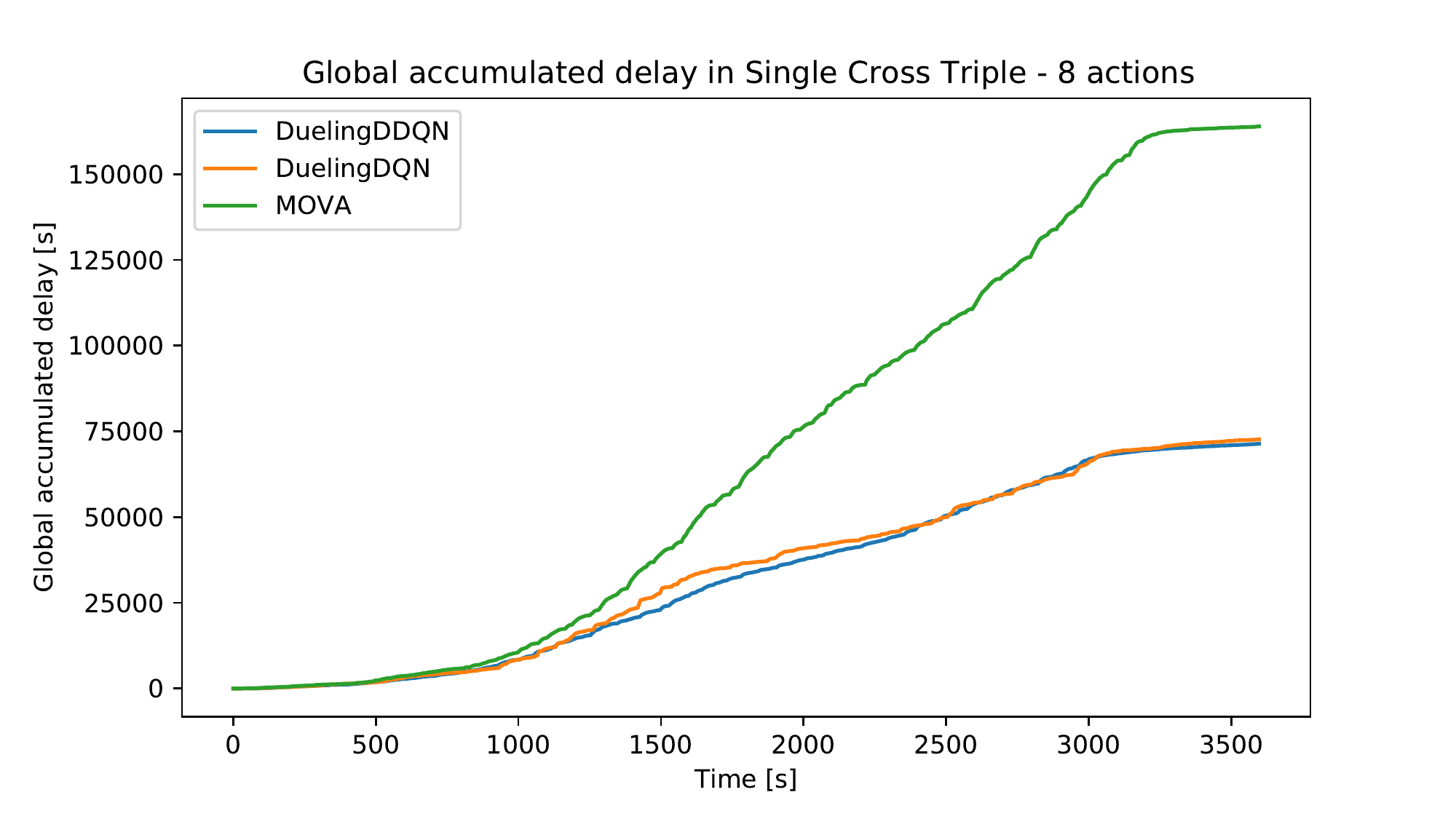}
  \caption{}
  \label{fig:scsdelaywo}
\end{subfigure}
\caption{(a) Global Cumulative Delay in Cross Triple - 4 actions. (b) Global Cumulative Delay in Cross Triple - 8 actions.}
\label{fig:scsresu}
\end{figure}
\begin{figure}
\begin{subfigure}{.45\textwidth}
  \centering
  \includegraphics[width=1\linewidth]{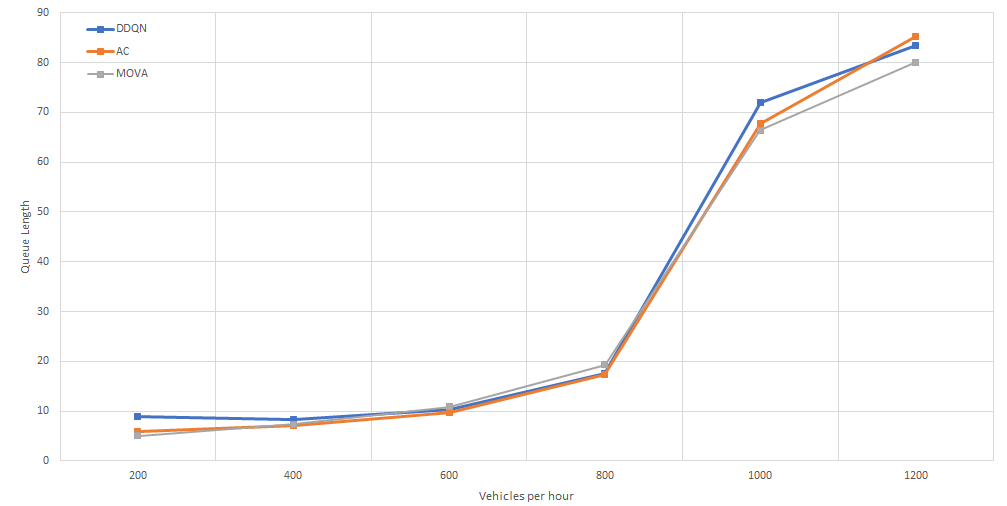}
  \caption{}
  \label{fig:scs length demand}
\end{subfigure}%
\begin{subfigure}{.45\textwidth}
  \centering
  \includegraphics[width=1\linewidth]{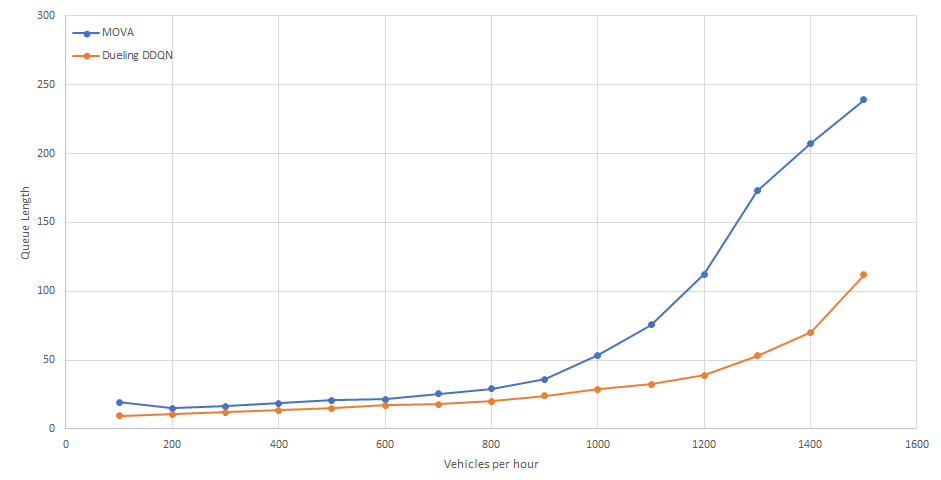}
  \caption{}
  \label{fig:scsdelaywo}
\end{subfigure}
\caption{(a) Queue length by demand level in Cross Straight. (b) Queue length by demand level in Cross Triple.}
\label{fig:scsresu}
\end{figure}
\begin{table}[htbp]
	\caption{Cumulative Delay in seconds and Average Sum of Max Queue length in meters on Cross Triple - 4 actions.}
	\begin{center}
		\begin{tabular}{|c|c|c|}
			\hline
			\textbf{Controller}&{\textbf{Cumulative Delay [s]}}&{\textbf{Average Max Queue Length [m]}} \\
			\hline
			MOVA& 260257.65 & 179.27\\
			\hline
			DDQN&  135220.91  & 153.58\\
			\hline
			DDDQN& 155563.22 & 128.20\\
			\hline
		\end{tabular}
		\label{table:queuesct4}
	\end{center}
\end{table}

\noindent \textbf{Experiment 3: Cross Triple - 8 actions. }
In order to allow the use of a higher variety of stages in the controllers the map was reworked.
All lanes were partitioned into their own independent links, allowing extra space for lane changes.
While these modifications allowed using information from all lanes in an akin manner to what modern sensors would achieve, due the lane changing limitations, direct comparisons with Experiment 2 must be handled with care. 
Both models share name and rough geometry, but the lanes layout is changed and so are the routing possibilities.

The results presented below, use DQN agents taking $12$ queue length inputs plus the state of the signal.
There are $8$ different stages. Here there are $4$ additional stages available for non-conflicting cross traffic. 
No specific stage order is enforced, and the agents are free to change between any combination of stages.

\begin{table}[htbp]
	\caption{Cumulative Delay and Cumulative Stop Delay in seconds across Controllers on Cross Triple - 8 actions.}
	\begin{center}
		\begin{tabular}{|c|c|c|}
			\hline
			\textbf{Controller}&{\textbf{Cumulative Delay [s]}}&{\textbf{Average Sum of Queues [m]}} \\
			\hline
			MOVA& 165456.44 & 339.41\\
			\hline
			DDQN& 72642.59 & 123.52\\
			\hline
			DDDQN& 71245.61 & 119.86\\
			\hline
		\end{tabular}
		\label{table:queuesct8}
	\end{center}
\end{table}
\vspace{-2mm}

The RL agents display a similar gap in performance with MOVA as in the previous experiment, with both classes benefiting from the increased actions pool. 
RL agents manage to generate about a third of the delay produced by MOVA.
While this appears to be a great success, these results have to be put into context.
MOVA has a lot of internal parameters meant to be fine tuned by a traffic engineer with site-specific knowledge.
Our settings did produce a successful control loop, operating in line with what was expected of the configuration process.
None of the RL agents has been fine tuned to the level that would be expected during commercial operation.
The layers and neuron distribution wasn't optimised, nor were the activation functions, meaning that the RL agents can still be improved upon.
\vspace{-2mm}
\section{Discussion}
\label{rlconclu}
\vspace{-1mm}
\noindent Several neural network architectures for RL controllers were tested.
The agents did not require extensive or complex configurations to adequately control traffic junctions, outperforming the commercial controllers.
RL agents showed great stability and robustness to control situations within their training envelope as well as outside of it.
Additionally, agents trained on relatively low uniform demand showed they can perform better than commercial systems during evaluation tests that included variable demand 5 times  higher than anything experienced during training.

In Experiment 1 MOVA and the RL agent following a DuelingDDQN architecture obtained very similar results, with a slight advantage for the RL agent.
Experiment 2 implies less granularity in the data and makes the control task more challenging. The results followed the same pattern with a DuelingDDQN agent obtaining the best performance over an array of loadings, despite the lower resolution in the input data.
Experiment 3 saw the introduction of a much more complex intersection.
Once again RL agents obtained significantly better results than MOVA, with the DuelingDDQN agent obtaining the lowest global and stop delay.
The gap between the performance of MOVA and RL agents is increased here with respect to the last experiment.
Most likely reasons are higher granularity in the data and extra actions being available to the agent, allowing it to display more complex sequences of actions.

We find that Reinforcement Learning applied to UTC can significantly outperform current adaptive traffic controllers in realistic commercial simulation software. 
These experiments provide credible evidence that Reinforcement Learning based UTC will be part of the next generation of traffic signal controllers.\\

\noindent \textbf{Acknowledgements.} This work was part funded by EPSRC Grant EP/L015374, by The Alan Turing Institute and the Toyota Mobility Foundation.
The authors thank Dr. W. Chernicoff for the discussions that made this project possible.

\bibliographystyle{elsarticle-harv}
\bibliography{CIT}

\begin{thebibliography}{21}
\expandafter\ifx\csname natexlab\endcsname\relax\def\natexlab#1{#1}\fi
\providecommand{\url}[1]{\texttt{#1}}
\providecommand{\href}[2]{#2}
\providecommand{\path}[1]{#1}
\providecommand{\DOIprefix}{doi:}
\providecommand{\ArXivprefix}{arXiv:}
\providecommand{\URLprefix}{URL: }
\providecommand{\Pubmedprefix}{pmid:}
\providecommand{\doi}[1]{\href{http://dx.doi.org/#1}{\path{#1}}}
\providecommand{\Pubmed}[1]{\href{pmid:#1}{\path{#1}}}
\providecommand{\bibinfo}[2]{#2}
\ifx\xfnm\relax \def\xfnm[#1]{\unskip,\space#1}\fi
\bibitem[{Bell and Bretherton(1986)}]{bell1986ageing}
\bibinfo{author}{Bell, M.C.}, \bibinfo{author}{Bretherton, R.D.},
  \bibinfo{year}{1986}.
\newblock \bibinfo{title}{{Ageing of fixed-time traffic signal plans}}, in:
  \bibinfo{booktitle}{International conference on road traffic control}.
\bibitem[{Cabrejas-Egea and Connaughton(2020)}]{multimodalarxiv}
\bibinfo{author}{Cabrejas-Egea, A.}, \bibinfo{author}{Connaughton, C.},
  \bibinfo{year}{2020}.
\newblock \bibinfo{title}{{Assessment of Reward Functions in Reinforcement
  Learning for Multi-Modal Urban Traffic Control under Real-World
  limitations}}.
\newblock \bibinfo{journal}{arXiv preprint arXiv:2010.08819}
  \href{http://arxiv.org/abs/arXiv:2010.08819}{{\tt arXiv:arXiv:2010.08819}}.
\bibitem[{{Cabrejas Egea} and Connaughton(2020)}]{ieeewarp}
\bibinfo{author}{{Cabrejas Egea}, A.}, \bibinfo{author}{Connaughton, C.},
  \bibinfo{year}{2020}.
\newblock \bibinfo{title}{{Wavelet Augmented Regression Profiling (WARP):
  improved long-term estimation of travel time series with recurrent
  congestion}}, in: \bibinfo{booktitle}{IEEE Conference on Intelligent
  Transportation Systems, Proceedings, ITSC}.
\newblock \DOIprefix\doi{10.1109/itsc45102.2020.9294318}.
\bibitem[{{Cabrejas Egea} et~al.(2018){Cabrejas Egea}, {De Ford} and
  Connaughton}]{egea2018estimating}
\bibinfo{author}{{Cabrejas Egea}, A.}, \bibinfo{author}{{De Ford}, P.},
  \bibinfo{author}{Connaughton, C.}, \bibinfo{year}{2018}.
\newblock \bibinfo{title}{{Estimating Baseline Travel Times for the UK
  Strategic Road Network}}, in: \bibinfo{booktitle}{IEEE Conference on
  Intelligent Transportation Systems, Proceedings, ITSC},
  \bibinfo{organization}{IEEE}. pp. \bibinfo{pages}{531--536}.
\newblock \DOIprefix\doi{10.1109/ITSC.2018.8569924}.
\bibitem[{{Cabrejas Egea} et~al.(){Cabrejas Egea}, Howell, Knutins and
  Connaughton}]{egearlutc}
\bibinfo{author}{{Cabrejas Egea}, A.}, \bibinfo{author}{Howell, S.},
  \bibinfo{author}{Knutins, M.}, \bibinfo{author}{Connaughton, C.}, .
\newblock \bibinfo{title}{{Assessment of Reward Functions for Reinforcement
  Learning Traffic Signal Control under Real-World Limitations}}, in:
  \bibinfo{booktitle}{IEEE International Conference on Systems, Man and
  Cybernetics, Proceedings, SMC}.
\newblock \DOIprefix\doi{10.1109/smc42975.2020.9283498}.
\bibitem[{Gao et~al.(2017)Gao, Shen, Liu, Ito and Shiratori}]{gao2017}
\bibinfo{author}{Gao, J.}, \bibinfo{author}{Shen, Y.}, \bibinfo{author}{Liu,
  J.}, \bibinfo{author}{Ito, M.}, \bibinfo{author}{Shiratori, N.},
  \bibinfo{year}{2017}.
\newblock \bibinfo{title}{{Adaptive traffic signal control: Deep reinforcement
  learning algorithm with experience replay and target network}}.
\newblock \bibinfo{journal}{arXiv preprint arXiv:1705.02755} .
\bibitem[{Genders and Razavi(2018)}]{genders2018}
\bibinfo{author}{Genders, W.}, \bibinfo{author}{Razavi, S.},
  \bibinfo{year}{2018}.
\newblock \bibinfo{title}{{Evaluating reinforcement learning state
  representations for adaptive traffic signal control}}.
\newblock \bibinfo{journal}{Procedia computer science} \bibinfo{volume}{130},
  \bibinfo{pages}{26--33}.
\bibitem[{Hasselt(2010)}]{hasselt2010double}
\bibinfo{author}{Hasselt, H.V.}, \bibinfo{year}{2010}.
\newblock \bibinfo{title}{{Double Q-learning}}, in:
  \bibinfo{booktitle}{Advances in neural information processing systems}, pp.
  \bibinfo{pages}{2613--2621}.
\bibitem[{Heydecker(2004)}]{heydecker2004}
\bibinfo{author}{Heydecker, B.G.}, \bibinfo{year}{2004}.
\newblock \bibinfo{title}{{Objectives, stimulus and feedback in signal control
  of road traffic}}.
\newblock \bibinfo{journal}{Journal of Intelligent Transportation Systems}
  \bibinfo{volume}{8}, \bibinfo{pages}{63--76}.
\bibitem[{Hunt et~al.(1982)Hunt, Robertson, Bretherton and Royle}]{SCOOT}
\bibinfo{author}{Hunt, P.B.}, \bibinfo{author}{Robertson, D.I.},
  \bibinfo{author}{Bretherton, R.D.}, \bibinfo{author}{Royle, M.C.},
  \bibinfo{year}{1982}.
\newblock \bibinfo{title}{{The SCOOT on-line traffic signal optimisation
  technique}}.
\newblock \bibinfo{journal}{Traffic Engineering {\&} Control}
  \bibinfo{volume}{23}.
\bibitem[{INRIX(2019)}]{inrix_scorecard_nodate}
\bibinfo{author}{INRIX}, \bibinfo{year}{2019}.
\newblock \bibinfo{title}{{Scorecard}}.
\newblock \URLprefix \url{http://inrix.com/scorecard/}.
\bibitem[{Liang et~al.(2018)Liang, Du, Wang and Han}]{liang2018}
\bibinfo{author}{Liang, X.}, \bibinfo{author}{Du, X.}, \bibinfo{author}{Wang,
  G.}, \bibinfo{author}{Han, Z.}, \bibinfo{year}{2018}.
\newblock \bibinfo{title}{{Deep Reinforcement Learning for Traffic Light
  Control in Vehicular Networks}} \bibinfo{volume}{XX}, \bibinfo{pages}{1--11}.
\newblock \URLprefix \url{http://arxiv.org/abs/1803.11115},
  \href{http://arxiv.org/abs/1803.11115}{{\tt arXiv:1803.11115}}.
\bibitem[{Mnih et~al.(2013)Mnih, Kavukcuoglu, Silver, Graves, Antonoglou,
  Wierstra and Riedmiller}]{mnih2013playing}
\bibinfo{author}{Mnih, V.}, \bibinfo{author}{Kavukcuoglu, K.},
  \bibinfo{author}{Silver, D.}, \bibinfo{author}{Graves, A.},
  \bibinfo{author}{Antonoglou, I.}, \bibinfo{author}{Wierstra, D.},
  \bibinfo{author}{Riedmiller, M.}, \bibinfo{year}{2013}.
\newblock \bibinfo{title}{{Playing atari with deep reinforcement learning}}.
\newblock \bibinfo{journal}{arXiv preprint arXiv:1312.5602} .
\bibitem[{Mousavi et~al.(2017)Mousavi, Schukat and Howley}]{mousavi2017}
\bibinfo{author}{Mousavi, S.S.}, \bibinfo{author}{Schukat, M.},
  \bibinfo{author}{Howley, E.}, \bibinfo{year}{2017}.
\newblock \bibinfo{title}{{Traffic light control using deep policy-gradient and
  value-function-based reinforcement learning}}.
\newblock \bibinfo{journal}{IET Intelligent Transport Systems}
  \bibinfo{volume}{11}, \bibinfo{pages}{417--423}.
\newblock \DOIprefix\doi{10.1049/iet-its.2017.0153}.
\bibitem[{Schaul et~al.(2016)Schaul, Quan, Antonoglou and
  Silver}]{schaul2015prioritized}
\bibinfo{author}{Schaul, T.}, \bibinfo{author}{Quan, J.},
  \bibinfo{author}{Antonoglou, I.}, \bibinfo{author}{Silver, D.},
  \bibinfo{year}{2016}.
\newblock \bibinfo{title}{{Prioritized experience replay}}.
\newblock \bibinfo{journal}{4th International Conference on Learning
  Representations, ICLR 2016 - Conference Track Proceedings}
  \href{http://arxiv.org/abs/1511.05952}{{\tt arXiv:1511.05952}}.
\bibitem[{Silver et~al.(2017)Silver, Schrittwieser, Simonyan, Antonoglou,
  Huang, Guez, Hubert, Baker, Lai, Bolton and Others}]{silver2017mastering}
\bibinfo{author}{Silver, D.}, \bibinfo{author}{Schrittwieser, J.},
  \bibinfo{author}{Simonyan, K.}, \bibinfo{author}{Antonoglou, I.},
  \bibinfo{author}{Huang, A.}, \bibinfo{author}{Guez, A.},
  \bibinfo{author}{Hubert, T.}, \bibinfo{author}{Baker, L.},
  \bibinfo{author}{Lai, M.}, \bibinfo{author}{Bolton, A.},
  \bibinfo{author}{Others}, \bibinfo{year}{2017}.
\newblock \bibinfo{title}{{Mastering the game of go without human knowledge}}.
\newblock \bibinfo{journal}{nature} \bibinfo{volume}{550},
  \bibinfo{pages}{354--359}.
\bibitem[{Smith and Barlow(2013)}]{SURTRAC}
\bibinfo{author}{Smith, S.F.}, \bibinfo{author}{Barlow, G.J.},
  \bibinfo{year}{2013}.
\newblock \bibinfo{title}{{SURTRAC : Scalable Urban Traffic Control}}.
\newblock \bibinfo{journal}{Transport Researach Board} .
\bibitem[{Stevanovic and Martin(2008)}]{stevanovic2008split}
\bibinfo{author}{Stevanovic, A.}, \bibinfo{author}{Martin, P.T.},
  \bibinfo{year}{2008}.
\newblock \bibinfo{title}{{Split-cycle offset optimization technique and
  coordinated actuated traffic control evaluated through microsimulation}}.
\newblock \bibinfo{journal}{Transportation Research Record}
  \bibinfo{volume}{2080}, \bibinfo{pages}{48--56}.
\bibitem[{Vincent and Peirce(1988)}]{MOVA}
\bibinfo{author}{Vincent, R.A.}, \bibinfo{author}{Peirce, J.R.},
  \bibinfo{year}{1988}.
\newblock \bibinfo{title}{{'MOVA': Traffic Responsive, Self-optimising Signal
  Control for Isolated Intersections}}.
\newblock \bibinfo{publisher}{Traffic Management Division, Traffic Group,
  Transport and Road Research}.
\bibitem[{Wan and Hwang(2018)}]{wan2018}
\bibinfo{author}{Wan, C.H.}, \bibinfo{author}{Hwang, M.C.},
  \bibinfo{year}{2018}.
\newblock \bibinfo{title}{{Value-based deep reinforcement learning for adaptive
  isolated intersection signal control}}.
\newblock \bibinfo{journal}{IET Intelligent Transport Systems}
  \bibinfo{volume}{12}, \bibinfo{pages}{1005--1010}.
\bibitem[{Xie et~al.(2012)Xie, Smith, Lu and Barlow}]{Xie2012}
\bibinfo{author}{Xie, X.F.}, \bibinfo{author}{Smith, S.F.},
  \bibinfo{author}{Lu, L.}, \bibinfo{author}{Barlow, G.J.},
  \bibinfo{year}{2012}.
\newblock \bibinfo{title}{{Schedule-driven intersection control}}.
\newblock \bibinfo{journal}{Transportation Research Part C: Emerging
  Technologies} \bibinfo{volume}{24}, \bibinfo{pages}{168--189}.
\newblock \DOIprefix\doi{10.1016/j.trc.2012.03.004}.

\end{thebibliography}
\end{document}